\documentclass{article}
\usepackage{proceedings}
\usepackage{times}
\usepackage{graphicx}
\usepackage{latexsym}
\usepackage{epsfig}

\newcommand{\E}{{\cal E}}

\newcommand{\SwitchOn}{\mbox{\textit{SwitchOn}}}
\newcommand{\SwitchOff}{\mbox{\textit{SwitchOff}}}
\newcommand{\Break}{\mbox{\textit{Break}}}
\newcommand{\Normal}{\mbox{\textit{Normal}}}
\newcommand{\Light}{\mbox{\textit{Light}}}

\begin{document}

\title{Modeling Complex Domains\\ of Actions and Change}

\author{Antonis Kakas and Loizos Michael\\
Dept. of Computer Science\\
University of Cyprus\\ P.O.Box 20537, CY-1678 Nicosia, Cyprus\\
antonis@ucy.ac.cy, cs98ml1@ucy.ac.cy}

\date{}

\maketitle
\bibliographystyle{ecai2002}

\begin{abstract}
This paper studies the problem of modeling complex domains of
actions and change within high-level action description
languages. We investigate two main issues of concern: (a) can we
represent complex domains that capture together different
problems such as ramifications, non-determinism and concurrency
of actions, at a high-level, close to the given natural ontology
of the problem domain and (b) what features of such a
representation can affect, and how, its computational behaviour.
The paper describes the main problems faced in this
representation task and presents the results of an empirical
study, carried out through a series of controlled experiments, to
analyze the computational performance of reasoning in these
representations. The experiments compare different
representations obtained, for example, by changing the basic
ontology of the domain or by varying the degree of use of
indirect effect laws through domain constraints. This study has
helped to expose the main sources of computational difficulty in
the reasoning and suggest some methodological guidelines for
representing complex domains. Although our work has been carried
out within one particular high-level description language, we
believe that the results, especially those that relate to the
problems of representation, are independent of the specific
modeling language.

\end{abstract}

\section{Introduction and Background}

Over the last decade several new frameworks for reasoning about
actions and change have been proposed. Together with the original
frameworks of the situation and event calculi, we have seen a
number of different action description languages emerging with
recent implemented prototype systems
\cite{Gelf98,LangL,LangC,KaMi97ais}. Many of these languages have
been motivated by the desire to have simple representation
frameworks, with semantics that relate more directly to the
natural intended meaning of their statements. They, therefore,
aspire to provide high-level representations of domains that are
close to their natural specification.

This paper studies the problem of modeling complex domains of
reasoning about actions and change within high-level action
description languages. It aims to expose the main problems
involved in this representation task and to study the
computational behaviour of the ensuing representations. It
considers the Zoo domain, proposed at the Logic Modeling Workshop
of the ETAI electronic communication forum \cite{LMW}, as one
representative complex domain that frameworks of reasoning about
actions and change should aim to capture.

A complex domain brings together a number of different problems
associated with reasoning about actions and change, such as
ramifications, non-determinism and concurrency. Although all
these aspects have been extensively studied and several solutions
have been proposed these studies are largely done for one problem
in isolation and with little attention to the computational
behaviour of these solutions. In this work we study how we can
use the knowhow generated over the years in these separate
problems, in order to represent complex and realistic domains
that encompass together all these different features. Despite the
fact that we will use a particular representation framework, that
of the Language $\E$, we believe that the results apply more
generally. This language encompasses (partial) solutions to the
main problems of reasoning about action and change that are based
on the same principles used by most of the other approaches. Hence
the findings of our work, especially those that refer to the
representational task would apply to other formalisms as well.

In particular, we investigate two main issues of concern: (a) can
we represent complex domains in a natural way using directly the
given natural ontology of the problem domain and (b)
what features of such a representation can affect (and how) its
computational behaviour. Our method of study uses an empirical
investigation through a series of controlled experiments. In some
of these experiments we compare different representations by
varying, for example, the basic ontology of the domain or the
degree of use of indirect effect laws through domain constraints.
In other experiments, we investigate the performance of a fixed
representation as we vary the type of problems we consider, e.g by
varying the degree of incompleteness of information in a scenario
or by varying the complexity of the narrative with irrelevant
action occurrences.

These experiments help to analyze the computational difficulties
of the reasoning by exposing their main sources. In turn, they
suggest methodological guidelines on how to develop complete
representations of complex domains with improved computational
properties.

In the rest of this section we present briefly the specific
action description language that we will use in our modeling
study. Section 2 studies the main problems faced in the task of
representing a complex domain (by analyzing these for the Zoo
domain). Section 3 discusses some of the experiments carried out
to study the computational behaviour of representations and
presents their main conclusions. Section 4 discusses the overall
conclusions drawn from our work.

\subsection{The Action Description Language $\E$}

Action description languages have been developed over the last
years to provide high level representation frameworks for theories
about actions and change. They generally do not have an explicit
representation of the frame axioms and they have been extended to
address the ramification and qualification problems. They have
been motivated by the need to have simple representation
frameworks with natural semantics that relate more directly to the
natural interpretations of the domains.

We will consider the problem of modeling complex domains within
the particular action description Language $\E$
\cite{KaMi97ais,KaMi97bis}. The vocabulary of the Language $\E$
consists of a set $\Phi$ of {\it fluent constants}, a set of {\it
action constants}, and a partially ordered set $\langle \Pi ,
\preceq \rangle$ of {\it time-points}. This vocabulary depends
each time on the domain being modeled. A {\it fluent literal} is
either a fluent constant $F$ or its negation $\neg F$. In the
current implementation of the language, the only time structure
that is supported is that of the natural numbers, so we restrict
our attention here mainly to domains of this type.

{\em Domain descriptions} in the Language ${\cal E}$ are
collections of the following types of statements (where $A$ is an
action constant, $T$ is a time-point, $F$ is a fluent constant,
$L$ is a fluent literal and  $C$ is a set of fluent literals):

\begin{itemize}
    \item  {\em t-propositions}
           of the form: $L \mbox{ {\bf holds-at} } T$

    \item  {\em h-propositions}
           of the form: $A \mbox{ {\bf happens-at} } T$

    \item  {\em c-propositions} of the form:
    $A \mbox{ {\bf initiates} } F  \mbox{ {\bf when} }C$
   or, $A \mbox{ {\bf terminates} } F  \mbox{ {\bf when} } C$

    \item  {\em r-propositions} of the form: $L  \mbox{ {\bf whenever} } C$

    \item  {\em p-propositions} of the form: $A  \mbox{ {\bf needs} }
    C$.

\end{itemize}
For the purposes of this paper it is sufficient to describe the
semantics of the language in an informal way, giving the intended
meaning of each type of sentence and describing the basic tenets
of the reasoning supported. T-propositions record observations
that particular fluents hold or do not hold at particular
time-points. H-propositions state that particular actions occur
at particular time-points. C-propositions state general ``action
laws'' -- the intended meaning of ``$A$ {\bf initiates} $F$ {\bf
when} $C$'' is that ``$C$ is a minimally sufficient set of
conditions for an occurrence of $A$ to initiate $F$''.
R-propositions, also called {\bf ramifications or domain
constraints}, serve a dual role in that they describe both static
constraints between fluents and ways in which fluents may be
indirectly affected by action occurrences. As a constraint this
must be satisfied as a classical implication at any time point.
The intended meaning of ``$L$ {\bf whenever} $C$'' is that ``at
every time-point that $C$ holds, $L$ holds, and hence every action
occurrence that brings about $C$ also brings about $L$''.
P-propositions state necessary conditions for an action to occur.

The semantics of ${\cal E}$ are based on a notion of a {\em model}
of a domain $D$ that assigns a truth value to each fluent at each
time point. A model encompasses two basic requirements:

\begin{description}
\item [Persistence] Direct and indirect action laws,
stated by c-propositions and through r-propositions respectively,
are the only way to bring about a change (over time) in the truth
value of a fluent, when an appropriate action occurs. This new
value persists in time from then onwards, until another such point
of change.

\item [Consistency] The model must satisfy as classical
implications the r-propositions at each time point and as facts
the t-propositions together with all the pre-conditions C at times
T that follow from each p-proposition $A  \mbox{ {\bf needs} } C$
and associated h-proposition $A \mbox{ {\bf happens-at} } T$.
\end{description}

Given a consistent domain $D$ and a query or goal $G$ comprising a
set of t-propositions, we say that $G$ is a {\bf credulous or
possible} conclusion of $D$ iff there is a model of $D$ in which
$G$ is true and that $G$ is a {\bf skeptical or necessary}
conclusion of $D$ iff $G$ is true in all models of $D$.

As an example, consider the following  simple ``bulb domain'' with
action constants $\SwitchOn$, $\SwitchOff$ and $\Break$ and
fluents $\Light$ and $\Normal$:\\

\noindent
$\mbox{\ \ \ }\SwitchOn \mbox{ {\bf initiates} } \Light \mbox{ {\bf when} } \{ \Normal \}$ \\
$\mbox{\ \ \ }\SwitchOff \mbox{ {\bf terminates} } \Light $ \\
$\mbox{\ \ \ }\Break \mbox{ {\bf terminates} } \Normal $ \\
$\mbox{\ \ \ }\neg \Light \mbox{ {\bf whenever} } \{\neg \Normal\}$ \\
$\mbox{\ \ \ }\SwitchOn \mbox{ {\bf needs} } \{\neg \Light\}$ \\
$\mbox{\ \ \ }\SwitchOn \mbox{ {\bf happens-at} } 2$ \\
$\mbox{\ \ \ }\Normal \mbox{ {\bf holds-at} } 0$ \\

\noindent This domain has as a skeptical conclusion $ \Light$ {\tt
holds-at} $4$, but the conclusion is only credulous when we remove
the last sentence.

The Language ${\cal E}$ has been implemented \cite{ERES} via an
argumentation-based translation of the language. This system,
called ${\cal E}$-RES, can support directly a variety of modes of
common sense reasoning such as: default persistence in credulous
or skeptical form, assimilation of observations and their
diagnosis possibly under incomplete information, as well as
combinations of these. The computational model on which this is
based integrates argumentation goal-oriented proofs for default
persistence, together with classical theorem proving techniques
for the satisfaction of the constraints expressed in a domain. It
also employs a (weak) notion of syntactic relevance in order to
help focus the computation to the relevant part of the theory.

This argumentation-based reformulation also extends the Language
${\cal E}$ allowing its action laws to be treated as default laws.

\section{Complex Domain Representations}

In this section we study how the Zoo domain, defined at the Logic
Modeling Workshop (LMW) \cite{LMW}, can be represented in the
Language ${\cal E}$. We will follow closely the presentation of
the domain and guidelines given at LMW trying to address all the
issues raised there. This section is therefore best read in
conjunction with the documentation available at LMW. We will
present the different aspects of representing such a complex
domain, while trying to expose the main problems faced and
choices made.

The LMW describes the Zoo domain as "...a world containing the
main ingredients of a classical zoo: cages, animals in the cages,
gates between two cages, as well as gates between a cage and the
exterior. In the Zoo world there are animals of several species,
including humans. Actions in the world may include movement
within and between cages, opening and closing gates, ..., riding
animals, etc."

\subsection{Background Landscape and Active Structure}

Our first task is to represent the {\bf landscape and active
structure} of the domain i.e. describe the background information
of the Zoo environment and the objects that populate it. This
information is generally taken to be static, not changing over
time.

For this type of information the Language ${\cal E}$ has a simple
facility of using {\bf constant} fluents and declaring the static
state of affairs for these at some initial time point (time 0).
For example, we declare the known animals in the domain by a set
of statements of the following form:

\noindent
$\mbox{ \ \ \ } animal(-) \ {\bf is\_constant}$\\
$\mbox{ \ \ \ } animal(john) \ \mbox{ {\bf holds-at} } 0$\\
$\mbox{ \ \ \ } animal(jane) \ \mbox{{\bf holds-at} }  0$\\
$\mbox{ \ \ \ } animal(elly) \ \mbox{ {\bf holds-at} } 0$\\
$\mbox{ \ \ \ }...$

In addition, we can use ramification statements to represent
static information that is derived from other such information
e.g. that ``an animal is large when it is an adult and belongs to
a large species'':

\noindent $\mbox{ \ \ \ } animal\_is\_large(A) \mbox{ {\bf
whenever} } \{animal\_is\_adult(A), \\  \mbox{ \ \ \ \ \ \ }
animal\_species(A,S), \ species\_is\_large(S) \}.$

In the same way, we represent the zoo terrain by declaring all
its positions and the neighbor relation via a constant fluent
$neighbor\_pos(\_,\_)$ and ramification statements to express its
properties e.g. $neighbor\_pos(P_1,P_2) \mbox{ {\bf whenever} }
\{neighbor\_pos(P_2,P_1) \}$ for the symmetry of this relation.

We then reason with the Closed World Assumption on these static
fluent predicates at time $0$ and project their truth value
unchanged to any other time point. If complete information for a
background fluent predicate is not known at $0$, then this
predicate would not be declared as constant and thus would be
interpreted under the Open World Assumption with the possibility
to assume its truth value.

We note that the inherent propositional nature of the Language
${\cal E}$ does pose some representational restrictions. We need
for example to define the primary typing constants, which
enumerate entities of the landscape and hence only domains with
known predefined entities can be represented. In general, the
underlying ontology of the language can be extended with a notion
of sorted variables over a set of finite sorts. The details of
this are not important for this work.

In this paper, we will assume that we have in the Zoo domain the
basic types of animals, species, gates, locations and positions,
defined as constant fluents, and that any variable (i=1,2,3,...)
denoted by $A_i, S_i, G_i, L_i, or \ P_i$ has the respective type.

\subsection{Action Effect Laws}

Till now we have only been concerned with the static aspects of
the representation of the domain i.e. representing information at
one time slice. The central part of any domain of reasoning about
actions and change is that of its {\bf Action laws specifying the
effects of actions}. In the Language ${\cal E}$ action laws are
represented directly via c-propositions. For example, the effect
of the $move\_to\_position(A,P)$ action in the Zoo domain, that
the animal ``A'' takes the position ``P'' is represented by:

\noindent $\mbox{ \ \ \ } \ move\_to\_position(A,P) \mbox{ {\bf
initiates} } animal\_pos(A,P) \\  \mbox{ \ \ \ \ \ \ } \mbox{ {\bf
when} } \{ reachable(A,P) \}$\footnote{Note that the current
version of the ${\cal E}$-RES system, that implements the Language
${\cal E}$, requires that the typing of each variable of a fluent
or action to be stated explicitly each time the fluent or action
is used in a statement. Hence in ${\cal E}$-RES this will written
as: $\mbox{ \ \ \ } \ move\_to\_position(A,P) \mbox{ {\bf
initiates} } animal\_pos(A,P) \\  \mbox{ \ \ \ \ \ \ } \mbox{
{\bf when} } \{animal(A), \ position(P), \ reachable(A,P) \}.$ }.

\noindent Here this effect law depends on the condition that at
the time of the action the position ``P'' is reachable by ``A'',
namely that ``P'' is an adjacent position to the position of the
animal ``A'' at the time of the action. This means that an
instance of the action $move\_to\_position(A,P)$ could occur
without successfully reaching ``P''. Alternatively, we can remove
this condition from the effect law and set it as a pre-condition
for the action to occur via the statement:

\noindent $\mbox{ \ \ \ } move\_to\_position(A,P) \mbox{ {\bf
needs} } \{reachable(A,P)\}.$

\noindent
 This choice depends on how we wish to interpret the
occurrence of events in a given narrative. If when given an event
we mean that this event has successfully occurred, then this
alternative is more appropriate and allows for more conclusions
to be drawn, given an action occurrence. For example, we would
include the statement

\noindent
 $\mbox{ \ \ } move\_to\_position(A,P) \mbox{ {\bf needs} }
\{ \neg rides(A,A_1)\}$

\noindent
 to capture the requirement that a rider cannot perform
the $move\_to\_position$ action at the time that s/he is riding.
Under this interpretation of event occurrences this way of
representing these preconditions allows us to draw more
conclusions from knowledge that an action has occurred. If then,
we are given that $move\_to\_position(john,p_1)$ happened at time
T then we would conclude that $john$ is not riding any (known)
animal at $T$. Depending on other information in the domain we
maybe able to project this backward or forward in time.

The above law of change for the action $move\_to\_position(A,P)$
expresses a {\bf direct} effect. This action can also have other
effects, for example that a rider $A_1$ of the animal $A$ will
also acquire the new position $P$, when $A$ moves to $P$. In
comparison, this is not a direct effect but nevertheless, as
argued in the recent literature (e.g. \cite{MichaelRams}),
actions should be allowed to have such {\bf indirect} effects.
This poses a dilemma in the representation. Indirect effects
could be represented either by explicit direct effect laws or via
the domain constraints (or ramification statements) of the theory.
In the above example we could have the additional direct laws:

\noindent $\mbox{ \ \ \ } \ move\_to\_position(A,P) \mbox{ {\bf
initiates} } animal\_pos(A_1,P) \\  \mbox{ \ \ \ \ \ \ } \mbox{
{\bf when} } \{
rides(A_1,A) \},$ \\
$ \mbox{ \ \ \ } move\_to\_position(A,P) \mbox{ {\bf terminates}
} animal\_pos(A,P_1) \\  \mbox{ \ \ \ \ \ \ } \mbox{ {\bf when} }
\{ animal\_pos(A,P_1) \}$

\noindent for moving an animal's rider along with the animal and
terminating the animal's current position, respectively. On the
other hand, we could omit these, since the following domain
constraints (stating that an animal's rider always has the same
position with the ridden animal and that an animal cannot be at
two positions at the same time) would generate the required
effects indirectly:

\noindent $\mbox{ \ \ \ } animal\_pos(A_1,P) \mbox{ {\bf whenever}
} \{animal\_pos(A,P), \\ \mbox{ \ \ \ \ \ \ } rides(A_1,A) \},$

\noindent $\mbox{ \ \ \ } \neg animal\_pos(A,P_1) \mbox{ {\bf
whenever} } \{animal\_pos(A,P), \\ \mbox{ \ \ \ \ \ \ } P_1 \neq
P \}.$

However, such domain constraints are also needed in the
representation to suitably restrict the assumptions that we might
make while reasoning, when we have incomplete information in the
theory.

Hence a dilemma emerges, either (1) to use direct laws to
explicitly enumerate all possible effects of an action together
with domain constraints which do not generate indirect effects,
but simply act as integrity constraints for assumptions relating
to incomplete fluents or (2) to use direct laws for the basic
effects of an action and let the domain ramification constraints
generate the other effects indirectly. We will see below that
these two choices can be semantically different and that they
differ in their computational performance.

\subsection{Domain Constraints}

To represent the whole Zoo domain we need to define all direct
action laws for each of the type of actions together with their
pre-conditions (as we have done above for $move\_to\_position$)
and state the various {\bf domain constraints} that the
specification of the problem requires. In addition, to the domain
constraints given above, we also have several other statements.
For example, we have:

\noindent $\mbox{ \ \ \ } false \mbox{ {\bf whenever} } \{
animal\_species(A,human),\\ \mbox{ \ \ \ \ \ \ }rides(A_1,A) \},$

 \noindent $\mbox{ \ \ \ } \neg rides(A,A_1) \mbox{ {\bf whenever} }
\{rides(A,A_2), A_1 \neq A_2 \}$

\noindent to capture the fact that a human cannot be ridden and to
represent the fact that an animal cannot ride two animals at the
same time. Here the first domain is written in this particular
way as a denial to indicate that it is not necessary to produce
indirect effects through it.

The domain constraints are also used to define various auxiliary
dynamic predicates that we need, such as $reachable$ used above:

\noindent $\mbox{ \ \ \ } reachable(A,P) \mbox{ {\bf whenever} }
\{animal\_pos(A,P_1), \\ \mbox{ \ \ \ \ \ \ } neighbor(P_1,P) \},$

\noindent $\mbox{ \ \ \ } \neg reachable(A,P) \mbox{ {\bf
whenever} } \{animal\_pos(A,P_1), \\ \mbox{ \ \ \ \ \ \ } \neg
neighbor(P_1,P) \}$

Note that here we do have the option (or dilemma) to define such
dynamic predicates via direct effect laws of the actions, but this
would be more complex and non-modular requiring the full
enumeration of all cases where the fluent $reachable$ could be
affected. Its definition through the above domain constraints and
their indirect effects that they embody, provides a compact and
modular way of representing such a dynamic predicate.

\subsection{Issues in the choice of Representation}

We return now to study in more detail the above dilemma and in
particular how this is related to non-determinism in the domain
and its representation. One of the actions of the Zoo domain, that
of $throwoff(A_1,A_2)$, is specified to be non-deterministic.
Animal $A_1$ throws off its back $A_2$ and the latter can land at
any position reachable from their common position at the time of
the throw. As there could be many such reachable positions and we
do not know at which one the animal would fall this is a
non-deterministic action. The argumentation semantics of the
Language ${\cal E}$ allows us  to represent such a
non-deterministic action law in the same way as any other law:

\noindent $\mbox{ \ \ \ }  throwoff(A_1,A_2) \mbox{ {\bf
initiates} } animal\_pos(A_2,P) \\ \mbox{ \ \ \ \ \ \ } \mbox{
{\bf when} } \{ reachable(A_2,P) \}.$

This effect law does not contain any explicit qualifications to
state that the animal would fall in one position only provided
that it does not fall on anyone of the other (reachable)
positions. This qualification is implicitly produced by the
domain constraint that an animal cannot have two different
positions at the same time. The combination of these two
statements captures the non-determinism of the action. An
inconsistency is not produced (although apparently it would be),
because for each direct initiation of a particular position $P$
that the action law produces, the domain constraint then produces
the indirect effect that all other positions are terminated. In
this way each position can be chosen. Note that it is important
that the blockage of the other positions is not captured simply
by the static application of the constraint, but that at the same
time the position of an animal changes to $P$ we also have a
change with regards to all the other positions, namely they
``begin to be false''.

We therefore see a useful need for the production of indirect
effects, i.e. of change through the domain constraints. This
provides a natural mechanism of implicit qualification of the
several direct laws for a non-deterministic action.
Alternatively, we would need to qualify explicitly the direct law
with the condition that the animal can not be at any other
position. This would be complex and difficult to express because
the qualifications to the action law need to refer to the state
of the world after the action has occurred.

The same behaviour can also be observed in the case of concurrent
actions whose effects are contradictory. Consider for example the
simultaneous occurrence of a $getoff(A,A_1,P)$ and a
$move\_to\_position(A_1,P_1)$ action, at a time when the animal
$A$ is riding another animal $A_1$. The first action requires
that $A$ has the position $P$ as its direct effect, while the
second action would give that $A$ has the different position
$P_1$ since $A$ is riding $A_1$. Again, the above constraint that
an animal can have only one position at a time, acts as a
qualifier and splits this scenario in a non-deterministic way into
two possibilities. In effect, in the first case $A$ manages to get
off before $A_1$ moves, whereas in the second case $A_1$ moves
first and then $A$ gets off. The domain constraint gives an
implicit qualification to the two effect laws involved, in a way
that would be very difficult (and unnatural) to represent
explicitly.

Concurrent action scenaria also show the main difference between
direct and indirect effect laws. Let us return to the indirect
effect law that a rider acquires the same position of the animal
that s/he is riding and compare it with its direct effect law
counterpart as given above. A difference between the two laws
shows up when we consider, for example, the concurrent $getoff$
and $move\_to\_position$ scenario, above. The direct law would
allow the rider to move to the new position with the animal,
whereas the indirect law would not generate this effect. This is
because the condition of $rides$ in the direct law case needs to
hold at the exact time of the action, while in the indirect law
case this needs to hold at an infinitesimal time (or next time
point) after this, so that both conditions in the ramification can
hold together to produce its conclusion. Indeed, due to the
simultaneous $getoff$ action, rides holds at the time of these
actions, but not later as it is terminated by this action
unconditionally. Hence, in this concurrent scenario we would not
get the second possibility of the indirect generation of the rider
moving with the animal.

In this way we see that direct laws can be stronger than a
corresponding indirect effect law and in fact could be used to
break non-determinism when we do not want this. To see this better
and show its significance for the task of modeling a domain, let
us consider a case where the non-determinism could exist as a
result of the occurrence of one action alone. Consider the domain
constraint that two animals at different positions cannot ride
each other:

\noindent $\mbox{ \ \ \ } \neg rides(A,A_1) \mbox{ {\bf whenever}
} \{animal\_pos(A,P), \\ \mbox{ \ \ \ \ \ \ }
animal\_pos(A_1,P_1),P_1 \neq P \}.$

This would then give indirect effects of the termination of riding
when we perform actions that change the position of the rider
alone (such as $getoff$ or $throwoff$). But it would also give the
indirect effect that riding is terminated when the animal moves
and hence it changes its position relative to its rider. In fact,
adding the above ramification to the representation and performing
an action $move\_to\_position(A_1,P_1)$, at a time point where the
animal $A_1$ is ridden by another animal, results in two
possibilities: either the rider comes off and is at the original
position of $A_1$, or the rider stays on and is at the new
position $P_1$ where the animal has moved to. With regards to the
effect of the action $move\_to\_position(A_1,P_1)$ on the position
of the rider the action is non-deterministic.

In this case we do not want this and we have a preference for the
rider to move along with the animal that it rides. We can achieve
this preference in a natural way by explicitly stating  the
preferred effect through a direct effect law. Despite the fact
that we have a domain constraint that gives, as an indirect
effect, that the rider moves along with the animal, we {\bf also}
represent this explicitly by a direct initiation law for the
action $move\_to\_position(A,P)$, i.e. we keep both sentences in
the dilemma presented above in this section. This then blocks the
indirect effect of terminating the riding by the movement of the
animal as the required condition for this, namely that the rider
is at a different position immediately after the action, cannot
now be supported due to the strong argument of changing its
position to that of the animal's new position through the new
direct law and the constraint that the rider can only have one
position at a time. The explicit statement of the direct law has
given preference to it.

In summary, we see that the ramifications and their indirect
effects provide an implicit form of qualification through the
general constraint properties of the domain. This means that the
theory treats its effect laws as default laws where one can
qualify another through these constraints. The direct effect laws
set out preferences amongst conflicting possibilities. The
redundancy of repeating an indirect law as a direct law as well,
forms a natural way of setting a preference. We give emphasis to
effects by articulating them explicitly.

\subsection{An example Scenario}

Closing this section, we show how a particular scenario can be
represented. A scenario is given by stating events, i.e. specific
instances of actions that occurred (through h-propositions) and
observations, that is values of fluents at specific time points
(through t-propositions). For example,

\noindent
$\mbox{ \ \ \ } throwoff(elly,john) \mbox{ {\bf happens-at} } 1$ \\
$\mbox{ \ \ \ } move\_to\_position(john,p1) \mbox{ {\bf happens-at} } 2$\\
$\mbox{ \ \ \ } mount\_animal(john,dumpo) \mbox{ {\bf happens-at} } 3$ \\
$\mbox{ \ \ \ } reachable(john,p1) \mbox{ {\bf holds-at} } 2$

\noindent
 represents the scenario that $elly$ threw off $john$, who
 then observed that position $p1$ was reachable, moved to position $p1$,
and then mounted $dumpo$. In this scenario we would reason to
$rides(john,elly)$ at time 1 or 0 as a necessary (or skeptical)
conclusion, but that $animal\_pos(john,p2)$ at time 2, where $p2$
is any neighbor of $p1$, is a possible (or credulous) conclusion.
At time 4 onwards we can conclude that $rides(john,dumpo)$ is
necessarily true. But if we are given also that
$move\_to\_position(dumpo,p3) \mbox{ {\bf happens-at} } 3$, then
$rides(john,dumpo)$ at 4 will only be a possible conclusion.
Observing that $animal\_pos(john,p3) \mbox{ {\bf holds-at} } 5$
i.e that $john$ and $dumpo$ have the same position then
$rides(john,dumpo)$ at 4 would be a necessary conclusion.

\section{Computational Behaviour of Representations}

In this section we study empirically, via suitable controlled
experiments, the computational behaviour of a representation and
try to ascertain what factors of the representation might or
might not affect this behaviour. We have conducted two major
types of experiments: (a) given a fixed representation of the Zoo
domain, we have examined various scenaria that differ in the
amount of given information, its relevancy to the queries and the
type of queries performed, and (b) given some fixed scenarios and
queries, we have examined various representations that differ on
the extend the action effects are represented directly/indirectly
and on the richness of the vocabulary used. The basic scale of
the domains on which we have carried these experiments is of the
order of 3 thousand ground clauses per time point. A scenario
typically spans over 3-6 time points.

As the representative results\footnote{All numbers shown in the
tables of this paper are in seconds measured on a Pentium II, 266
MHz, PC. The absolute value of these times are not important as
the purpose of the experiments is to compare different cases.}
presented in figure 1 suggest, increasing the information in a
scenario can affect  positively the computational performance,
even if the extra information given explicitly in the scenario is
a necessary conclusion of this.
\begin{figure}[h]
\centerline{\resizebox{8truecm}{!}{\epsffile{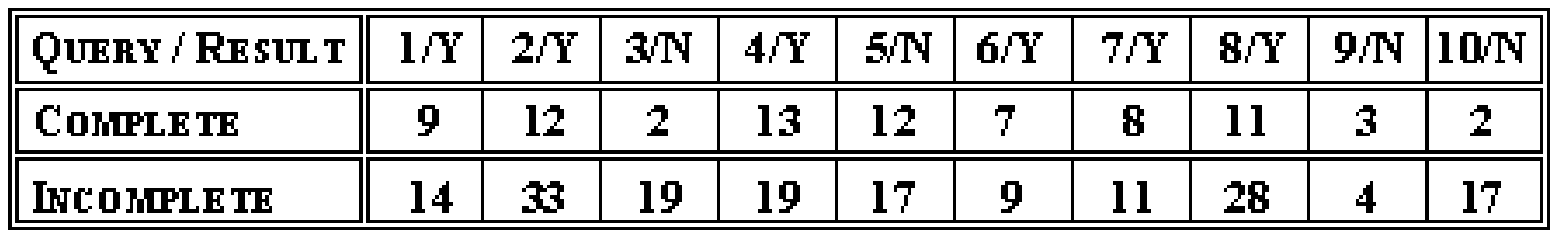}}}
\caption{Complete vs Incomplete Scenario Knowledge}
\label{FigTable1}
\end{figure}

These experiments have also shown that it is possible to use
simple syntactic relevancy checks to focus the computation to the
relevant (w.r.t. the query at hand) part of the domain. In fact,
the experiments have helped us develop further a semantically
enhanced notion of relevancy that neutralizes to some extent the
effect of additional irrelevant information in the scenario.
Figure 2 shows that the addition of irrelevant actions in a
scenario does not have a significant impact on the computation.
The stability under irrelevant information also depends strongly
on the type of representation used and the choice of vocabulary,
as we will see below in the second class of experiments. Finally,
as expected in some cases, particularly in scenarios with a lot of
missing information, skeptical queries can be significantly
slower than the corresponding credulous queries.

\begin{figure}[h]
\centerline{\resizebox{8truecm}{!}{\epsffile{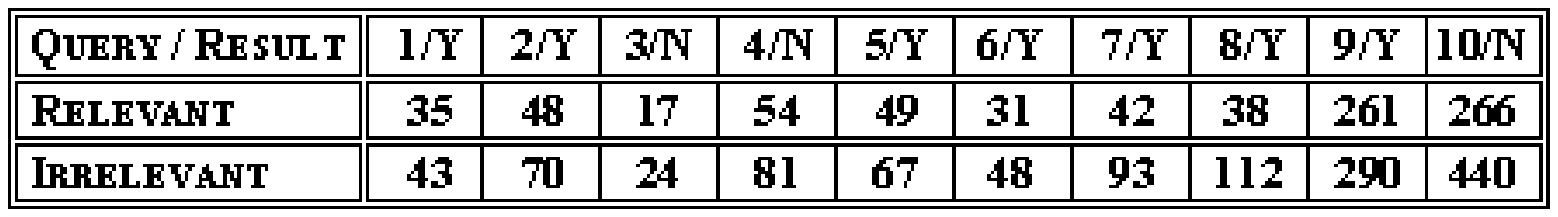}}}
\caption{Relevant vs Additional Irrelevant Actions}
\label{FigTable2}
\end{figure}

As mentioned above, the second class of experiments considers
changes in the representation. Figure 3 compares the
computational behaviour between representations which use direct
effect laws to a different degree. In  general, using only direct
laws gives a significantly better computational performance than
a representation that uses to a maximum indirect effect laws.  We
should note though, as we have seen in section 2, that the first
representation cannot always fully capture the domain e.g. the
non-determinism of conflicting concurrent actions. In such cases,
this becomes inconsistent and is unable to return an answer. It
is also noteworthy, that when effect laws are duplicated in the
representation both directly and indirectly, the behaviour varies
from being as good as a representation that primarily uses direct
effects, to being worse than a representation that primarily uses
indirect effects. This depends mostly on whether the specific laws
that are duplicated as direct laws can generate a large number of
additional indirect effects.

\begin{figure}[h]
\centerline{\resizebox{8truecm}{!}{\epsffile{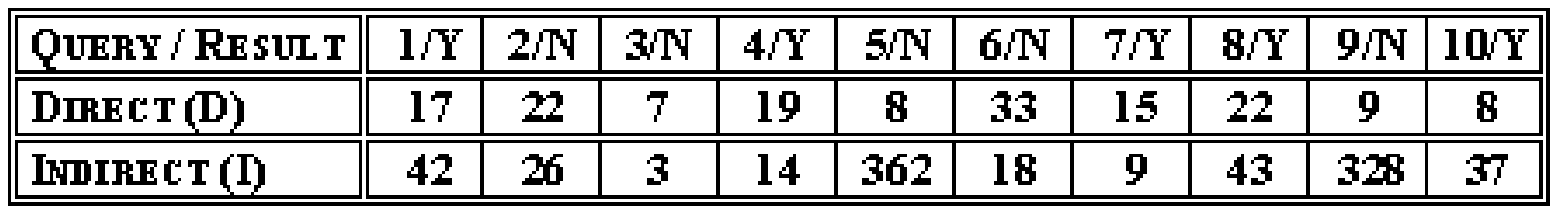}}}
\caption{Direct vs Indirect Effect Laws} \label{FigTable2}
\end{figure}

Another series of experiments has examined the effects on the
computational behaviour of using an enriched, more specific
vocabulary of the domain in the representation. Starting from a
minimally sufficient vocabulary, we have developed a number of
other equivalent representations using more specialized
vocabularies that separate explicitly  different specific cases
for an action, e.g. the single action fluent $move\_to\_position$
is replaced by $ move\_in\_cage$, $enter\_cage$ and $exit\_cage$.
Similarly, we can distinguish explicitly in the vocabulary some
of the fluent properties e.g. those that refer to the (2
dimensional) position on the floor from the properties that refer
to the third dimension off the floor. We have found that a
representation that uses, in this way, a carefully selected
ontology and proper typing can significantly improve its
computational efficiency. This occurs mainly by reducing the
number of assumptions (sometimes impossible ones) from being
considered in the reasoning.

With regards to the issue of scaling of the computational
behaviour of the representations, our experiments suggest that the
representation that maximizes the use of direct effect laws
scales up relatively well. These are preliminary results, tested
only on small sized domains that contain up to a maximum of 15
positions in the terrain, corresponding to around 25,000 ground
clauses. Representations with non-minimal use of indirect effect
laws scale up poorly.

In general, our empirical study has revealed two main
computational bottlenecks: (a) the number of additional
consequences drawn from the conclusions of a query that need to
be considered to ensure that the query can be consistently
satisfied in the whole domain and (b) the number of
counter-proofs (counter-arguments) that are generated via
persistence on an assumption needed to prove the query. Our
experiments have helped us address (at least partly) these
difficulties. The enhanced notion of relevancy, which helps us
avoid unnecessary re-computation of known consequences and a
(non-minimal) choice of vocabulary that distinguishes explicitly
different scenaria, can result in significant improvement of
computational behaviour. Another potential source of improvement
can come from the way we use direct and indirect effect laws.
Generally, the more use of direct laws the better the
performance, but there is a trade off here with the reduction in
modularity and readability of a representation that uses direct
laws extensively. In some cases the computational gain is
relatively small to justify this and a representation that
combines the use of  these two types of effect laws is overall
better.

We have also begun to examine the same problems using a different
computational model in place of the argumentation based model
used so far. We have translated the Language $\E$ to a SAT theory
so that we can then employ a SAT solver for query answering. This
translation is not complete and supports only a restricted class
of domains. In particular, it does not support domains with
non-deterministic actions, or concurrent actions that generate
conflicting effects nor domains in which the ramifications form
cycles. Initial results indicate that the SAT-based approach has
a more stable behaviour. They also indicate that it scales up
better as we increase the number of actions in a scenario, but not
so as we increase the number of positions in the terrain of the
Zoo domain. An interesting difference of the SAT-based approach is
the fact that it is difficult to employ relevancy notions in it.
Some simple forms of relevancy can be used, but this requires
that for each query we need to translate the relevant part of the
domain theory afresh and this can have an adverse effect on its
performance.

\section{Related Work and Conclusions}

We have studied how to model complex domains of action and change
based on the theoretical foundations that the community at large
has developed over the last decade. In particular, we have
addressed the challenge set by the Logic Modeling Workshop
\cite{LMW}, to apply this theory in modeling complex domains that
encompass together many different aspects of the problem. As in
the related work of \cite{traffic,ZooCCalc,TrafficCCalc} our work
shows that the existing frameworks of action description
languages are sufficiently expressive to meet this
representational challenge.

Moreover, applying this otherwise well understood theory of
reasoning about actions and change to capture complex domains has
helped to expose more clearly some problems. For example, it has
shown a link between non-determinism and an implicit
qualification of effect laws via the domain constraints where the
explicit representation of an indirect effect law as a direct law
can give this higher priority. As mentioned in the introduction,
we believe that these considerations are independent of the
particular framework that we have used to carry out our study and
that they would apply also to other approaches that are based on
the same underlying principles for developing a theory for
reasoning about actions and change. Differences may arise in the
computational aspects of the different frameworks. For this
reason we have concentrated more on comparison experiments and
have tried to exposed inherent computational problems. Also the
propositional nature of the particular framework that we have
used could be limiting in the static aspects of the
representation, i.e. in the ability to represent knowledge at any
single time or situation. But this does not affect the ability to
represent the dynamic aspects of a problem domain. At this
initial stage of the study we believe that it is methodologically
correct to de-couple these two aspects and examine at a later
stage how they would affect each other.

Our study of the computational behaviour of representations has
suggested some methodological guidelines for modeling complex
domains that can help us control their computational performance.
These include the balanced use of direct and indirect effect laws
and the adoption of an appropriate vocabulary as presented in the
previous section. To cross check these comparison results it would
be useful to study similar experiments using other languages, e.g.
using the Causal Calculator \cite{causalcalc} with its SAT based
computation. We also need to study how other mechanisms, external
to the reasoning, can help focus more the computation to the
relevant part of the theory.

More work is needed to understand better what factors can
influence the computational performance of a representation and
how this would scale up. We also need to develop a stricter
methodology on how to choose a vocabulary for a complex domain
that would give it good computational properties. In our work so
far this choice was mostly motivated by the natural common sense
reasoning that we as humans apply to the domain and our use of a
specific vocabulary for it. In fact, a distinction between
``natural'' and ``formal'' (or compact) representations emerges,
where the first type of representation uses a non-minimal
enriched vocabulary which helps it encode directly in the domain
qualitative solutions to computational problems (e.g. the
geometry of a terrain) and thus results to a better computational
behaviour. This and its possible link to qualitative reasoning in
AI is an interesting problem for future study.

Our work has been mainly motivated with trying to understand a
natural computational model of human common-sense reasoning and
how this is affected both by issues of representation and methods
of computation. In the future we plan to investigate, through a
series of cognitive experiments, how natural are these
representations and how well they capture the human common sense
reasoning in the domain of action and change.

\subsubsection*{Acknowledgements}

This work was partly supported by the European Union KIT project
CLSFA (9621109) and IST project SOCS (IST-2001-32530).

\bibliography{NMR02-camera-1}

\begin{thebibliography}{10}

\bibitem{LangL}
M.~Gelfond C.~Baral and A.~Provetti, `Representing actions: Laws, observations
  and hypotheses', in {\em JLP 31(1-3), pp. 201-243}, (1997).

\bibitem{LMW}
ETAI Forum, `http://www.ida.liu.se/ext/etai/lmw'.

\bibitem{Gelf98}
M.~Gelfond and V.~Lifschitz, `Action languages', in {\em ETAI}, volume 3(16),
  (1998).

\bibitem{LangC}
E.~Giuchiglia and V.~Lifschitz, `An action language based on casual
  explanation', in {\em AAAI-98}, pp. 623--630, (1998).

\bibitem{traffic}
A.~Hensechel and M.~Thielscher, `Logic modeling workshop: The traffic world',
  in {\em ETAI}, volume 3, nr 7, http://www.ida.liu.se/ea/cis/1999/007, (1999).

\bibitem{ZooCCalc}
V.~Lifschitz J.~Lee and H.~Turner, `A representation of the zoo world in the
  language of the causal calculator', in {\em Proc. Common Sense 2001}.

\bibitem{KaMi97ais}
A.C. Kakas and R.S. Miller, `A simple declarative language for describing
  narratives with actions', in {\em JLP 31(1-3), pp. 157-200}, (1997).

\bibitem{KaMi97bis}
A.C. Kakas and R.S. Miller, `Reasoning about actions, narratives and
  ramifications', in {\em ETAI Vol 1(4), pp. 39-72}, (1998).

\bibitem{ERES}
A.C. Kakas, R.S. Miller, and F.~Toni, `E-res: Reasoning about actions, events
  and observations', in {\em LPNMR'01, pp. 254-266}, (2001).

\bibitem{causalcalc}
N.~McCain, {\em Causality in Commonsense Reasoning about Actions}, PhD thesis,
  University of Texas at Austin, 1997.

\bibitem{MichaelRams}
M.~Thielscher, `Ramification and causality', in {\em Artificial Intelligence},
  volume 89 (1-2), pp. 317-364, (1997).

\bibitem{TrafficCCalc}
J.~Lee V.~Akman, S. T.~Erdogan and V.~Lifschitz, `A representation of the
  traffic world in the language of the causal calculator', in {\em Proc. Common
  Sense 2001}, (2001).

\end{thebibliography}

\end{document}